%% file: Graph-Guided.tex
\documentclass{article} 
\usepackage{iclr2021_conference,times}

\input{math_commands.tex}

\usepackage{hyperref}
\usepackage{url}

\usepackage{amsfonts}
\usepackage{amsmath}
\usepackage{natbib}

\usepackage{booktabs}
\usepackage{graphicx}
\usepackage{subcaption}
\usepackage{bigstrut}
\usepackage{marvosym}
\usepackage{algorithm}
\usepackage{algorithmic}
\usepackage{listings} 
\usepackage{setspace}
\usepackage{microtype}
\usepackage{mathtools}

\title{A Graph-Guided Reasoning Approach for Open-ended Commonsense Question Answering}

\author{Zhen Han\thanks{\, work done while internship at Baidu.}, \; Yue Feng$^{1}$, \; Mingming Sun$^{1}$ \\
$^{1}$Cognitive Computing Lab, Baidu Research $\;$  \\
sunmingming01@baidu.com}

\definecolor{capri}{rgb}{0.0, 0.75, 1.0}

\iclrfinalcopy 
\begin{document}

\maketitle

\begin{abstract}
Recently, end-to-end trained models for multiple-choice commonsense question answering (QA) have delivered promising results. However, such question-answering systems cannot be directly applied in real-world scenarios where answer candidates are not provided. Hence, a new benchmark challenge set for open-ended commonsense reasoning (OpenCSR) has been recently released, which  contains natural science questions without any predefined choices. 
On the OpenCSR challenge set, many questions require implicit multi-hop reasoning and have a large decision space, reflecting the difficult nature of this task. Existing work on OpenCSR sorely focuses on improving the retrieval process, which extracts relevant factual sentences from a textual knowledge base, leaving the important and non-trivial reasoning task outside the scope. 
In this work, we extend the scope to include a reasoner that constructs a question-dependent open knowledge graph based on retrieved supporting facts and employs a sequential subgraph reasoning process to predict the answer. The subgraph itself can be seen as a concise and compact graphical explanation of the prediction. 
Experiments on two OpenCSR datasets show that the proposed model achieves great performance on benchmark OpenCSR datasets.

\end{abstract}

\section{Introduction}
Commonsense reasoning has long been considered an essential topic in artificial intelligence. Most approaches work on the setting of \textit{multiple-choice question answering} \citep{lin-etal-2019-kagnet, feng-etal-2020-scalable}, which selects an answer choice by scoring the question-choice pairs. However, the multiple-choice setting is not applicable in many real-world scenarios since many question-answering tasks do not provide answer candidates. As a step towards making commonsense reasoning research more realistic and useful, open-ended commonsense reasoning (OpenCSR) has been introduced \citep{lin2020differentiable}, which explores a commonsense knowledge corpus to answer commonsense questions. OpenCSR often requires multi-hop reasoning, i.e., the model should conclude the answer by reasoning over two or more facts from the knowledge corpus, which makes this task much more challenging. Lin et al. \citep{lin2020differentiable} proposed a retrieval-based method, called \textit{DrFact}, by combining the maximum inner product
search and symbolic links between facts. However, DrFact does not put much effort on the reasoning module to re-rank the retrieved facts. 
To this end, we proposed an integrated subgraph reasoning approach for OpenCSR with end-to-end learning, which iteratively employs a retriever to extract question-relevant facts from a knowledge corpus and a reasoner over the extracted facts. Given a commonsense question, the proposed approach applies DPR \citep{karpukhin-etal-2020-dense} to extract relevant facts from a textual knowledge corpus,
converts the retrieved natural language facts into a graph-structured format using Open Information Annotation (OIA) \cite{sun2020predicate} and performs subgraph reasoning on the constructed joint OIA-graph using a multi-relational graph attention network. Specifically, the reasoner first performs entity linking from the giving question to the joint OIA-graph. Then it starts from the linked entities (nodes), and iteratively samples relevant edges with a pruning procedure to form an enclosing subgraph around the question. 
The reasoning procedure takes into account both structural information, i.e., graph structure of the joint OIA graph, and semantic information, i.e., language representation of questions and facts. After several rounds of retrieval and pruning, the model predicts the answer from the concepts in the subgraph.

Our contributions are as follows: (1) we investigate how to perform a cooperative retrieval-and-reasoning in open-ended commonsense question answering. To the best of our knowledge, our work is the first  retrieve-and-reasoning approach for OpenCSR. (2) We present experimental results that
show our model achieves great results on the benchmark OpenCSR dataset with an ablation study demonstrating the 
performance gain of integration structural information and semantic information. (3) The proposed method can potentially homogenize structured, i.e., knowledge base, and unstructured commonsense knowledge, i.e., textual corpus for answering open-ended commonsense questions since it can unify both knowledge formats into a graph-structured format.  

\begin{figure*}
\begin{subfigure}{0.35\textwidth}
 \centering
   \includegraphics[width=\linewidth]{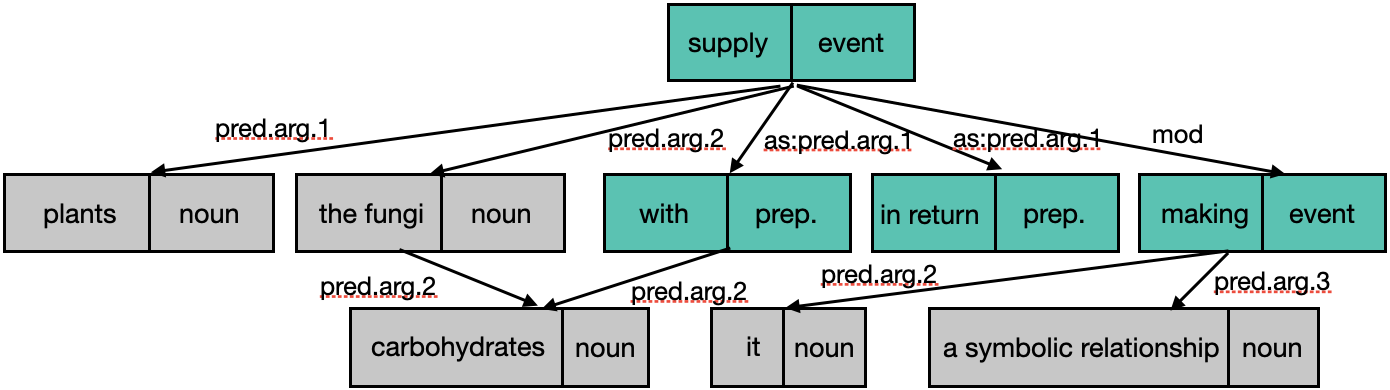}
   \caption{\label{fig: single oia}}
\end{subfigure}%
\begin{subfigure}{0.6\textwidth}
 \centering
   \includegraphics[width=\linewidth]{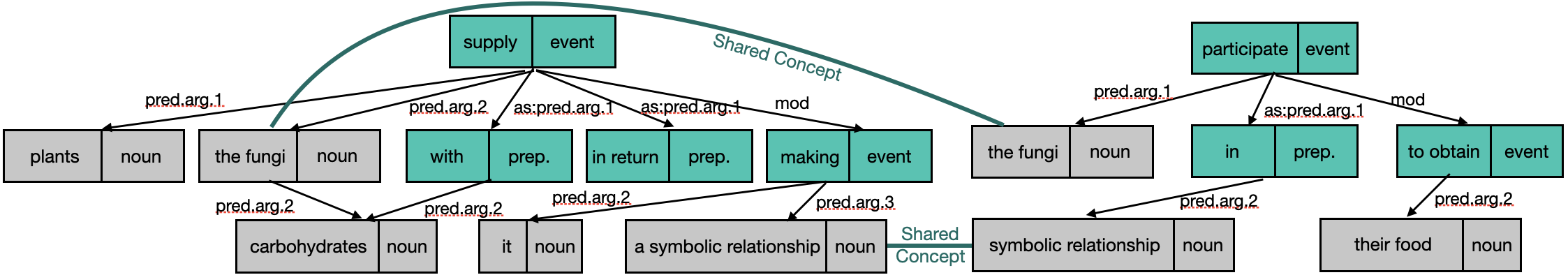}
   \caption{\label{fig: joint oia}}
\end{subfigure}%
 \caption{(\textbf{a}) The OIA graph of "Plants supply the fungi with carbohydrates, in return, making it a symbiotic relationship." There are two types of nodes: constant and predicate. Constant nodes are simple nominal phrases while predicate nodes include simple verbal phrases and prepositional phrases. Edges in OIA graphs are labeled. \textit{pred.arg.n} denotes the n-th arguments of a predicate node, \textit{mod} indicates the modification, and \textit{as:pred.arg.n} expresses an reversed relation of \textit{pred.arg.n}. (\textbf{b}) The joint OIA graph consists of two factoid sentences that share the concepts "fungi" and "symbiotic relationship". }
\end{figure*}

\section{Related Work}
\paragraph{Commonsense Reasoning}
Traditional commonsense reasoning (CSR) techniques are mainly designed for multiple-choice QA. For instance, to independently score each decision, KagNet \citep{lin-etal-2019-kagnet} and MHGRN \citep{feng-etal-2020-scalable} both leverage external commonsense knowledge graphs as structural priors. Although effective in selecting the best response for a multiple-choice question, these techniques are less useful for real-world situations because answer candidates are frequently unavailable. By fine-tuning a text-to-text transformer, UnifiedQA \citep{khashabi-etal-2020-unifiedqa}  
generated answers to questions. However, a drawback of multiple-choice QA models is that they do not provide intermediate explanations for their answers, making them less suitable in many real-world scenarios. Lin et al. \citep{lin2020differentiable} introduced the open-ended commonsense reasoning and proposed DrFact to directly retrieve relevant facts, and then use the concepts mentioned in the top-ranked facts as answer predictions. 

\paragraph{Subgraph Reasoning}
Many recent works learn representations of localized subgraphs. Alsentzer et al. \citep{alsentzer2020subgraph} introduced a subgraph neural network to learn disentangled subgraph representations using a novel subgraph routing mechanism. Teru et al. \citep{teru2020inductive} proposed a graph neural network that reasons over local subgraph structures and performs inductive relation predictions. Han et al. \citep{han2020explainable} developed an explainable reasoning framework for forecasting future links on temporal knowledge graphs by employing a sequential reasoning process over local subgraphs.

\begin{figure*}
 \centering
   \includegraphics[width=.82\linewidth]{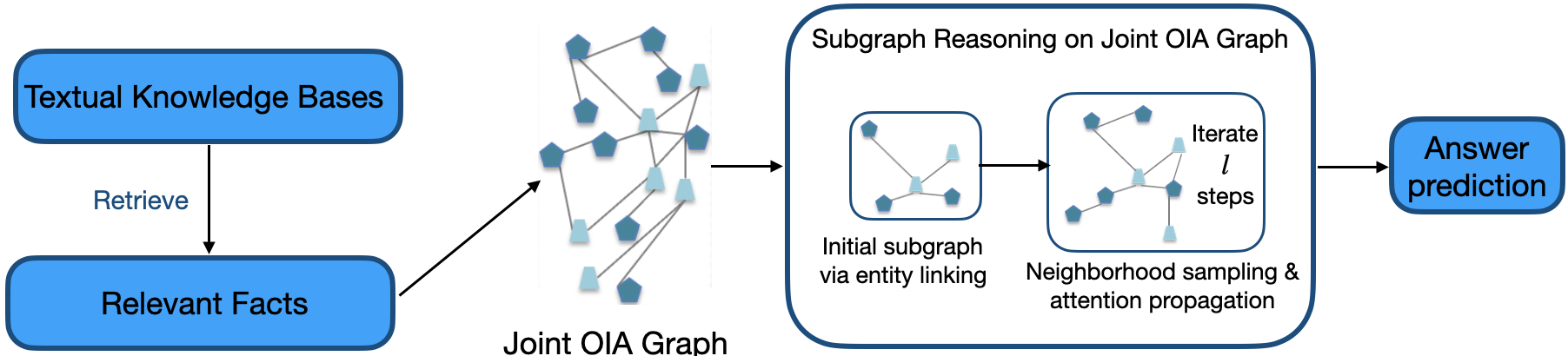}
   \caption{\label{fig:Framework Architecture} Model Architecture.}
\end{figure*}
   
\section{Our Approach}
\paragraph{Retrieving Relevant Facts}
Following the dense passage retrieval work \citep{karpukhin-etal-2020-dense}, we use a bi-encoder transformer architecture that learns to maximize the inner product of the representation of a question and the relevant factual sentences from the knowledge corpus containing correct answers to the given question. 

\paragraph{Constructing Question-dependent Joint OIA-Graph}
Following the steps in \citep{sun2020predicate}, we convert each retrieved factual sentence into an OIA-graph as shown in Figure \ref{fig: single oia}. For each node in an OIA graph, we link it with nodes in the OIA graphs of other sentences that include the same concept. We label this kind of link as \textit{shared concepts}. As shown in Figure \ref{fig: joint oia}, the factoid sentences "Plants supply the fungi with carbohydrates, in return, making it a symbiotic relationship." and "Fungi participate in symbiotic relationships to obtain their food." shares the same concepts "fungi" and "symbiotic relationship". 
Then, we construct a joint OIA-graph $\mathcal{G}_{joint}$ by linking nodes that share the same concepts in different OIA graphs.

\paragraph{Subgraph Reasoning on the Joint OIA-Graph}
Inspired by \cite{han2020xerte}, we conduct reasoning on a dynamically expanded inference graph $\mathcal G_{\textit{inf}}$ extracted from the joint OIA-graph. Given a commonsense question $q$, we build an initial inference graph via \textbf{entity linking} between the question $q$ and the joint OIA-graph. We find all nodes of $\mathcal{G}_{joint}$ that share the same concepts as $q$ includes. We set such OIA-nodes to be the initial nodes of the inference graph $\mathcal G_{\textit{inf}}$. The inference graph expands by sampling one-hop neighbors of initial nodes in $\mathcal{G}_{joint}$. Besides, we propose a semantic-following operation to build skip connections between the initial nodes and their multi-hop neighbors. Taking a node $v$ in $\mathcal G_{\textit{inf}}$ as an example, we compute the inner-product similarity between its representation and the representation of other nodes in $\mathcal{G}_{joint}$ obtained by the retrieval and add the top $K$ nodes into $\mathcal G_{\textit{inf}}$ by linking them with $v$. The contribution of the semantic-following has two folds: 1) It speeds up the reasoning process and broadens the receptive field of the subgraph reasoner by adding skip connections between multi-hop neighbors; 2) It allows the subgraph reasoner to take into account both semantic-relevant and symbolic-linked nodes regarding a given question. Next, we feed $\mathcal G_{\textit{inf}}$ into a relational graph attention layer that takes node embedding as the input, computes an attention score for each edge indicating the relevance to the given question, and produces a question-dependent representation for each node using message passing. Instead of treating all neighbors with equal importance in the massage passing, we take the question information into account and assign varying importance levels to each neighbor by calculating the following question-dependent attention score: 

\begin{equation} 
\begin{aligned}
\label{equa: attention score} 
e_{vu}^{l} (q, p_k) =\mathbf{W}_{s}^{l}(\mathbf{h}_v^{l-1} || \mathbf p_k^{l-1} || \mathbf h_{q}^{l-1})^T
\\ \mathbf{W}_{t}^{l}(\mathbf{h}_u^{l-1} || \mathbf p_k^{l-1} || \mathbf h_{q}^{l-1}),
\end{aligned}
\end{equation}
where $e_{vu}^{l} (q, p_k)$ is the attention score of the edge $(v, p_k, u)$ regarding the question $q$, $p_k$ corresponds to the edge type between the source node $v$ and the target node $u$, $\mathbf{W}_{s}^{l}$ and $\mathbf{W}_{t}^{l}$ are two weight matrices for capturing the dependencies between question representations and source node features specified for source node and target node, respectively. $\mathbf p_k$ is the edge embedding indicating the relationship between $u$ and $v$. $\mathbf h_v^{l-1}$ denotes the hidden representation of the node $v$ at the $(l-1)^{th}$ inference step. When $l=1$, i.e., for the first layer,  $\mathbf h_v^{0}$ is the aggregated token representation from Bert. Then, we compute the normalized attention score $\alpha_{vu}^{l} (q, p_k)$ using the \textit{softmax function}. 
Once obtained, we aggregate the representations of the sampled neighbors of node $v$ denoted as $\hat{\mathcal N}_v$ and weight them using the normalized attention scores, which are written as
\begin{equation}
\label{equa: hidden representation update}
\mathbf{h}^l_{v} (q) =\sum_{u \in \hat{\mathcal N}_v} \alpha_{vu}^{l} (q, p_k)\mathbf{h}^{l-1}_{u}(q).
\end{equation}

\begin{table*}[htpb]
    \centering
    \resizebox{.7\textwidth}{!}{
    \begin{tabular}{l|cccc|cccc|}
        \toprule
        Datasets & \multicolumn{4}{|c|}{\textbf{ARC-Open}} & \multicolumn{4}{|c|}{\textbf{OBQA-Open}}\\
         \midrule
        Model & H@50 & H@100 & R@50 & R@100 & H@50 & H@100 & R@50 & R@100\\
        \midrule\relax
         DPR & 68.67 & 78.62 & 28.93 & 38.63 & 54.47 & 67.73 & 15.17 & 22.34\\
         DrKIT & 67.63 & 77.89 & 27.57 & 37.29 & 61.74 & 75.92 & 18.18 & 27.10\\
         DrFact & 71.60 & 80.38 & 31.48 & 40.93 & 69.01 & 80.03 & 21.27 & 30.32\\
         Our model & 72.76 & 80.38 & 31.09 & 40.24 & 62.30 & 73.80 & 18.11 & 26.83\\
         \bottomrule
    \end{tabular}}
    \caption{Results of the Hit@K and Rec@K (K=50/100) in \% on OpenCSR.}\label{tab: prediction results}
\end{table*}

\paragraph{Answer Prediction}
We compute the plausibility score $s_{v,q}^l$ of node $v$ to be the answer of question $q$ at the $l^{th}$ inference step as follows:
\begin{equation}
s_{v,q}^l = s_{mips}(\mathbf q, \mathbf f_v) + \sum_{u\in \overline{\hat{\mathcal{N}}}_v} \sum_{p_k \in \mathcal P_{uv}}\alpha_{uv}^l(q, p_k) a_{u,q}^{l-1}, \label{equa: node score spread}
\end{equation}
where $s_{mips}(\mathbf q, \mathbf f_v)$ denotes the relevance score of the retrieved fact $f_v$, which mentions the node $v$, regarding the question $q$ by the maximum inner product search. Since the same concept may appear in different nodes in the inference graph, we aggregate the plausibility score of nodes that share the same concept to assign each concept a unique attention score: 
\begin{equation} 
s_{c_i, q}^l = g({s_{v,q}^l|v(c) = c_i}), \;\;\;\textrm{for}\, v\in \mathcal V_{\mathcal G_{\textit{inf}}},
\end{equation}
where $s_{c_i, q}^l$ denotes the plausibility score of concept $c_i$, $\mathcal V_{\mathcal G_{\textit{inf}}}$ is the set of nodes in inference graph $\mathcal G_{\textit{inf}}$. $v(c)$ represents the concept included in node $v$, and $g(\cdot)$ represents a score aggregation function.  Here we use the maximum function. 

\paragraph{Inference Graph Expansion and Pruning}
After several iterations of expansion, the inference graph $\mathcal G_{\textit{inf}}$ would grow rapidly and cover almost all nodes. To prevent the inference graph from exploding, we reduce the graph size by pruning the edges with a small plausibility score and keeping the edges with $K$ largest contribution scores. 
After running $L$ inference steps, the model selects the concept with the highest plausibility score in $\mathcal G_{\textit{inf}}$ as the answer to the given question, where the inference graph itself serves as a graphical explanation.

\paragraph{Loss Function}
We use the binary cross-entropy as the loss function, which is 
\begin{equation*}
\label{equa: loss function}
\begin{aligned}
\mathcal L & =  - \frac{1}{|\mathcal Q|}\sum_{q\in \mathcal Q}\frac{1}{|\mathcal C^{inf}_q|}\sum_{c_i \in \mathcal C^{inf}_q} 
(y_{c_i, q}\log(\frac{s_{c_i,q}^{L}}{\sum_{c_j \in \mathcal C^{inf}_q} s_{c_j,q}^{L}}) \\ &+ (1- y_{c_i, q}) \log((1-\frac{s_{c_i, q}^{L}}{\sum_{c_j \in \mathcal C^{inf}_q}s_{c_j,q}^{L}}))),
\end{aligned}
\end{equation*}
where $\mathcal C_q^{inf}$ represents the set of concepts in the inference graph of the question $q$, $y_{c_i, q}$ represents the binary label that indicates whether $c_i$ is an answer for $q$, and $Q$ denotes the question set. $s_{c_i,q}^{L}$ denotes the plausibility score of concept $c_i$ at the final inference step. 

\section{Experiments}
\paragraph{Fact corpus and concept vocabulary}
Following settings in \cite{lin2020differentiable}, GenericsKB-Best corpus serves as the main commonsense knowledge source that contains 1,025,413 unique facts. All sentences in the corpus are provided with concepts, which are frequent noun chunks, using the spaCy toolkit. There are 80,524 concepts in total.

\paragraph{Datasets and evaluation metrics} 
We evaluate our model on two benchmark open-ended commonsense reasoning datasets, i.e., ARC-Open and OBQA-Open \cite{lin2020differentiable}, that contain 6600 and 5288 questions, separately. Every question could be answered using various concepts, where the average answer is 6.8 and 7.7 in ARC-Open and OBQA-Open. Each dataset provides the set of true answer concepts for each question. We use two metrics, Hits@K and Recall@K, where Hits@K denotes the percentage of times that at least one true concept appears in the top k of ranked concepts.

\paragraph{Experimental Results}
\label{sec: ablation study}
We compare our model with DPR \cite{karpukhin-etal-2020-dense}, DrKIT \cite{dhingra2020differentiable}, and DrFact \cite{lin2020differentiable}. Recall that our model applies DPR as the retriever so it is a straightforward baseline. And DrFact is the strongest baseline in OpenCSR. As shown in Table \ref{tab: prediction results}, our model outperforms DPR and DrKIT on ARC-Open and achieves on-par performance as DrFact. All results are averaged over three trials. We provide implementation details in Appendix \ref{app: implementation} and attach the source code in the supplementary material.

\begin{table}[htpb]
    \centering
    \resizebox{.5\linewidth}{!}{
    \begin{tabular}{l|cccc|}
        \toprule
        Datasets & \multicolumn{4}{|c|}{\textbf{ARC-Open}}\\
         \midrule
        Model & H@50 & H@100 & R@50 & R@100\\
        \midrule\relax
         Model w/ SC & 72.76 & 80.38 & 31.09 & 40.24 \\
         Model w/o SC & 71.74 & 79.65 & 30.56 & 39.75 \\
         \bottomrule
    \end{tabular}}
    \caption{Ablation Study on ARC-Open: we investigate the gain of adding skip connections (SC) to semantically relevant multi-hop neighbors.}\label{tab: ablation study}
\end{table}

\paragraph{Ablation Study}
Recall that the proposed subgraph reasoner takes into account both the structurally linked one-hop neighbors and semantically relevant multi-hop neighbors by expanding the inference graph $\mathcal G_{inf}$. Table \ref{tab: ablation study} shows an ablation study in that we disable the reasoner to add semantically relevant multi-hop neighbors while inference graph expansion called \textit{Model w/o SC}, demonstrating the performance gain of integrating both structural and semantic information.

\section{Conclusion}
 We present a novel graph-guided neural symbolic commonsense reasoning approach for the open-ended commonsense reasoning task. The proposed method takes advantage of the dense passage retrieval and graph neural network reasoner to answer open-ended commonsense questions. Specifically, the graph reasoner integrates both structural dependency information between facts and semantic information by constructing an open information annotation graph and employing a semantic-following operation. The proposed model generates an inference graph for each question, which can be seen as a concise and compact graphical explanation of the prediction. The model achieved great performance on two benchmark datasets while being more interpretable.

\bibliography{iclr2021_conference}
\bibliographystyle{iclr2021_conference}

\appendix
\section*{Appendices}

\section{Limitations}
The proposed model performs a sequential reasoning process, and thus, may cause long inference time when answering requires a quite long multi-hop reasoning chain. Besides, the "share-link", which connects nodes that share the same concept, would have a significant amount in some datasets and make the underlying graph much denser. It would make it difficult for the model to decide the expansion direction of the subgraph. 

\section{The license of the Artifacts}
The datasets we used in this work are proposed by Lin et al. \cite{lin2020differentiable} which is licensed under the MIT License. 

About the model we proposed in this work, we will release it and give the license, copyright information, and terms of use once the paper gets accepted. 

\section{Implementation}
\label{app: implementation}
We tune the hyperparameters of our models using the random search and report the best configuration in the source code in the supplementary material. The training costs 73 GPU hours on the ARC-Open dataset and 50 GPU hours on the OBQA-Open dataset with NVIDIA A40 instance.

\end{document}

%% file: math_commands.tex
\usepackage{amsmath,amsfonts,bm}

\def\eqref#1{equation~\ref{#1}}

\def\1{\bm{1}}

\DeclareMathAlphabet{\mathsfit}{\encodingdefault}{\sfdefault}{m}{sl}
\SetMathAlphabet{\mathsfit}{bold}{\encodingdefault}{\sfdefault}{bx}{n}













%% file: Graph-Guided.bbl
\begin{thebibliography}{11}
\providecommand{\natexlab}[1]{#1}
\providecommand{\url}[1]{\texttt{#1}}
\expandafter\ifx\csname urlstyle\endcsname\relax
  \providecommand{\doi}[1]{doi: #1}\else
  \providecommand{\doi}{doi: \begingroup \urlstyle{rm}\Url}\fi

\bibitem[Alsentzer et~al.(2020)Alsentzer, Finlayson, Li, and
  Zitnik]{alsentzer2020subgraph}
Emily Alsentzer, Samuel Finlayson, Michelle Li, and Marinka Zitnik.
\newblock Subgraph neural networks.
\newblock \emph{Advances in Neural Information Processing Systems},
  33:\penalty0 8017--8029, 2020.

\bibitem[Dhingra et~al.(2020)Dhingra, Zaheer, Balachandran, Neubig,
  Salakhutdinov, and Cohen]{dhingra2020differentiable}
Bhuwan Dhingra, Manzil Zaheer, Vidhisha Balachandran, Graham Neubig, Ruslan
  Salakhutdinov, and William~W Cohen.
\newblock Differentiable reasoning over a virtual knowledge base.
\newblock \emph{arXiv preprint arXiv:2002.10640}, 2020.

\bibitem[Feng et~al.(2020)Feng, Chen, Lin, Wang, Yan, and
  Ren]{feng-etal-2020-scalable}
Yanlin Feng, Xinyue Chen, Bill~Yuchen Lin, Peifeng Wang, Jun Yan, and Xiang
  Ren.
\newblock Scalable multi-hop relational reasoning for knowledge-aware question
  answering.
\newblock In \emph{Proceedings of the 2020 Conference on Empirical Methods in
  Natural Language Processing (EMNLP)}, pp.\  1295--1309, Online, November
  2020. Association for Computational Linguistics.
\newblock \doi{10.18653/v1/2020.emnlp-main.99}.
\newblock URL \url{https://aclanthology.org/2020.emnlp-main.99}.

\bibitem[Han et~al.(2020{\natexlab{a}})Han, Chen, Ma, and
  Tresp]{han2020explainable}
Zhen Han, Peng Chen, Yunpu Ma, and Volker Tresp.
\newblock Explainable subgraph reasoning for forecasting on temporal knowledge
  graphs.
\newblock In \emph{International Conference on Learning Representations},
  2020{\natexlab{a}}.

\bibitem[Han et~al.(2020{\natexlab{b}})Han, Chen, Ma, and Tresp]{han2020xerte}
Zhen Han, Peng Chen, Yunpu Ma, and Volker Tresp.
\newblock xerte: Explainable reasoning on temporal knowledge graphs for
  forecasting future links.
\newblock \emph{arXiv preprint arXiv:2012.15537}, 2020{\natexlab{b}}.

\bibitem[Karpukhin et~al.(2020)Karpukhin, Oguz, Min, Lewis, Wu, Edunov, Chen,
  and Yih]{karpukhin-etal-2020-dense}
Vladimir Karpukhin, Barlas Oguz, Sewon Min, Patrick Lewis, Ledell Wu, Sergey
  Edunov, Danqi Chen, and Wen-tau Yih.
\newblock Dense passage retrieval for open-domain question answering.
\newblock In \emph{Proceedings of the 2020 Conference on Empirical Methods in
  Natural Language Processing (EMNLP)}, pp.\  6769--6781, Online, November
  2020. Association for Computational Linguistics.
\newblock \doi{10.18653/v1/2020.emnlp-main.550}.
\newblock URL \url{https://aclanthology.org/2020.emnlp-main.550}.

\bibitem[Khashabi et~al.(2020)Khashabi, Min, Khot, Sabharwal, Tafjord, Clark,
  and Hajishirzi]{khashabi-etal-2020-unifiedqa}
Daniel Khashabi, Sewon Min, Tushar Khot, Ashish Sabharwal, Oyvind Tafjord,
  Peter Clark, and Hannaneh Hajishirzi.
\newblock {UNIFIEDQA}: Crossing format boundaries with a single {QA} system.
\newblock In \emph{Findings of the Association for Computational Linguistics:
  EMNLP 2020}, pp.\  1896--1907, Online, November 2020. Association for
  Computational Linguistics.
\newblock \doi{10.18653/v1/2020.findings-emnlp.171}.
\newblock URL \url{https://aclanthology.org/2020.findings-emnlp.171}.

\bibitem[Lin et~al.(2019)Lin, Chen, Chen, and Ren]{lin-etal-2019-kagnet}
Bill~Yuchen Lin, Xinyue Chen, Jamin Chen, and Xiang Ren.
\newblock {K}ag{N}et: Knowledge-aware graph networks for commonsense reasoning.
\newblock In \emph{Proceedings of the 2019 Conference on Empirical Methods in
  Natural Language Processing and the 9th International Joint Conference on
  Natural Language Processing (EMNLP-IJCNLP)}, pp.\  2829--2839, Hong Kong,
  China, November 2019. Association for Computational Linguistics.
\newblock \doi{10.18653/v1/D19-1282}.
\newblock URL \url{https://aclanthology.org/D19-1282}.

\bibitem[Lin et~al.(2020)Lin, Sun, Dhingra, Zaheer, Ren, and
  Cohen]{lin2020differentiable}
Bill~Yuchen Lin, Haitian Sun, Bhuwan Dhingra, Manzil Zaheer, Xiang Ren, and
  William~W Cohen.
\newblock Differentiable open-ended commonsense reasoning.
\newblock \emph{arXiv preprint arXiv:2010.14439}, 2020.

\bibitem[Sun et~al.(2020)Sun, Hua, Liu, Wang, Zheng, and Li]{sun2020predicate}
Mingming Sun, Wenyue Hua, Zoey Liu, Xin Wang, Kangjie Zheng, and Ping Li.
\newblock A predicate-function-argument annotation of natural language for
  open-domain information expression.
\newblock In \emph{Proceedings of the 2020 Conference on Empirical Methods in
  Natural Language Processing (EMNLP)}, pp.\  2140--2150, 2020.

\bibitem[Teru et~al.(2020)Teru, Denis, and Hamilton]{teru2020inductive}
Komal Teru, Etienne Denis, and Will Hamilton.
\newblock Inductive relation prediction by subgraph reasoning.
\newblock In \emph{International Conference on Machine Learning}, pp.\
  9448--9457. PMLR, 2020.

\end{thebibliography}
